\documentclass[preprint,12pt]{elsarticle}%

\usepackage{natbib}
\usepackage[utf8]{inputenc}
\usepackage{graphicx}
\usepackage{amsmath}
\usepackage{amsfonts}
\usepackage{amssymb}
\usepackage{multirow}
\usepackage{array}
\usepackage{booktabs}
\usepackage{cancel}
\usepackage[pdftex]{hyperref}
\usepackage{amsthm}
\usepackage{multicol}
\usepackage{enumerate} 
\usepackage{color}
\hypersetup{colorlinks=true,allcolors=black} 
\usepackage[dvipsnames,svgnames,x11names,hyperref,table]{xcolor}
\usepackage{epsfig}
\usepackage{array}
\usepackage{colortbl}
\usepackage{algorithm}
\usepackage[noend]{algpseudocode}

\makeatletter
\def\ps@pprintTitle{%
 \let\@oddhead\@empty
 \let\@evenhead\@empty
 \def\@oddfoot{\footnotesize\itshape
       Postprint submitted to \ifx\@journal\@empty Elsevier
       \else\@journal\fi\hfill}
 \let\@evenfoot\@oddfoot}
\makeatother
\journal{Engineering Applications of Artificial Intelligence}

\begin{document}

\begin{frontmatter}



\title{Capturing waste collection planning expert knowledge in a fitness function through preference learning}


\author[arcelor]{Laura Fernández Díaz}
\ead{laura.fernandezdiaz@arcelormittal.com}
\author[arcelor]{Miriam Fernández Díaz}
\ead{miriam.fernandezdiaz@arcelormittal.com}
\author[uniovi]{Jos\'e Ram\'on Quevedo}
\ead{quevedo@uniovi.es}
\author[uniovi]{Elena Monta\~n\'es\corref{cor1}}
\ead{montaneselena@uniovi.es}

\cortext[cor1]{Corresponding author}

\address[uniovi]{Artificial Intelligence Center. University of Oviedo at Gij\'on, 33204 Asturias,  Spain \texttt{http://www.aic.uniovi.es}}

\address[arcelor]{Global R\&D, ArcelorMittal (Spain)}

\begin{abstract}
This paper copes with the COGERSA waste collection process. Up to now, experts have been manually designed the process using a trial and error mechanism. This process is not globally optimized, since it has been progressively and locally built as council demands appear. Planning optimization algorithms usually solve it, but they need a fitness function to evaluate a route planning quality.
The drawback is that even experts are not able to propose one in a straightforward way due to the complexity of the process. Hence, the goal of this paper is to build a fitness function though a preference framework, taking advantage of the available expert knowledge and expertise. Several key performance indicators together with preference judgments are carefully established according to the experts for learning a promising fitness function. Particularly, the additivity property of them makes the task be much more affordable, since it allows to work with routes rather than with route plannings. Besides, a feature selection analysis is performed over such indicators, since the experts suspect of a potential existing (but unknown) redundancy among them. The experiment results confirm this hypothesis, since the best $C-$index ($98\%$ against around $94\%$) is reached when 6 or 8 out of 21 indicators are taken. Particularly, truck load seems to be a highly promising key performance indicator, together to the travelled distance along non-main roads. A comparison with other existing approaches shows that the proposed method clearly outperforms them, since the $C-$index goes from $72\%$ or $90\%$ to $98\%$.
\end{abstract}
\end{frontmatter}
\section{Introduction}\label{sec:0}
Until now, each local council has been requested waste collection services to the company \textit{Consorcio para la Gestión de los Residuos Sólidos de Asturias} (COGERSA). The different requests have been appearing at different moments in time and every time a request has arisen, the experts of this company have just focused on a particular area to design specific routes, keeping unaltered the already planned routes. Hence, the actual routes have been locally optimized in each area object of request and one isolated from the rest. But now, the company requires unifying the waste collection planning in order to optimize their human and material resources, the collection time, fuel consumption and so on. The task is not as easy as it could seem because several issues must be taken into account. On one hand, there are several planning constraints imposed by the governments of each council. For instance, a container located at the entrance of the schools must not be picked up at the same time that the classes start or end. Also, the waste collection cannot be carried out when there is a street market planned. In addition, governments impose the collection frequencies and the number of containers. Also, the working time regulations must be considered, since, for instance, a same worker or truck cannot exceed a certain number of work hours. On the other hand, there are some other constraints about available resources. For instance, in regard to the laterality\footnote{Laterality is the property that has a container in order to indicate the side of the road to pick it up.}, the containers can be located to the right or to the left of the road, then, in case of two way roads, trucks must turn around  in order to be able to pick up the container depending on the direction of traffic. Another issue to take into account is the distance that a truck travels with its cargo box loaded, since it might be better for the truck to pick up the containers on the way back to travel less distance with the cargo box full.
\\
This problem is typically solved using planning optimization algorithms. The application of these techniques requires a fitness function able to evaluate the quality of a route planning. The main drawback is there exists a lack of a suitable function of this kind, and, even more, it is neither obvious nor easy to obtain in the waste collection. Hence, and firstly we need to face to the hurdle of designing a fitness function to evaluate a route planning, which will be the main focus and the main contribution of this paper. The plan in the future is to include this fitness function into an optimization algorithm scheme, but now this task is out of the scope of this paper. Despite the experts are not able to design a straightforward fitness function, they have a lot of expertise knowledge susceptible of being mined. Within artificial intelligence field, there exist two main ways of carrying out this practice. On one hand, if the experts are able to describe their knowledge, then this knowledge can be developed using rules leading to an expert system, for instance, for refrigerant flow air conditioning systems \cite{guo2019expert}. On the other hand, if only the actions of the experts are available but they are not able to explain them, then, it is possible to use machine learning techniques to extract their knowledge from those actions, for instance, for automated cyber security data triage \cite{zhong2018learning}. The latter is the case of the COGERSA Company, since experts perform actions from their expertise, but they are not able to display the reasons about their decisions. There exist several ways of taking into account the expert knowledge to feed a machine learning approach. A straightforward possibility could be to ask the experts for providing a score for a route planning and then perform a learning process through a regression system. However, this kind of approaches are not adequate in general, since experts ratings cannot be interpreted as absolute assessments. The main reason is that experts tend to rate in a relative way, hence, the same expert might provide a different score to the same route planning depending on if this route planning is evaluated in a context with much more better route plannings or if it is evaluated in a context with much more worse route plannings. Thus, the route planning presented in a batch surrounded by worse route plannings will probably get a higher score than if this route planning were showed together with better route plannings. This situation is known as batch effect and it often biases the ratings. Hence, feeding a regression method with this unstable data can mislead the learning process \cite{diez2004discovering}. However, it is possible to state that the expert assessments are consistent in regard to the order, that is, experts may provide the same order to the same route plannings set despite the score granted may differ from one expert to another. This evidence justifies that feeding a machine learning system with expert information in form of preference judgments is quite more promising than doing so in form of ratings, since avoiding the assumption that certain rating means the same for every expert avoids to mislead the learning process \cite{buck2001predicting, cohen99learning}. This is the reason why the proposal of this paper advocates to fit the problem of finding an adequate fitness function to assess a route planning into a preference learning framework \cite{bahamonde2007howtolearn}. This environment not only requires to capture expert knowledge and expertise in form of preference judgements, otherwise the route plannings have to be described through indicators experts must establish in order to machine learning takes place. Those indicators will be the features of the machine learning process and they must capture the key issues experts implicitly pay attention to when they establish their preference judgements. The reality is that both kind of information were not available, hence, experts were asked for both preference judgments and key performance indicators. In fact, one of the contributions of this paper is an exhaustive design of key performance indicators to establish the route planning descriptions. Another contribution is to design route planning pairs from the actual route planning to make experts show their preferences, for which a greedy algorithm will be designed. The third contribution is the preference learning itself. Finally, experts reported a highly existing redundancy among the key indicators. At this respect, they manually proposed to take several sets to explore their impact. However, this paper goes further and the fourth contribution consists of carrying out an automatic feature selection procedure to establish the relevance of the key performance indicators according to a machine learning procedure. There are other approaches in the literature that provide a fitness function from key indicators. For instance, different key performance indicators from mobile networks \cite{leontiadis2017good} are combined using two approaches. The first one consists of a regression procedure, for which absolute ratings are required, a task almost unachievable by the experts. Besides, and, as commented before, this approach suffers from batch effect that can mislead the learning process. Even more, a regression procedure is not able to optimize the ranking \cite{leontiadis2017good}. The second approach optimizes the Spearman's rank correlation using a particle swarm optimization, which is a population-based metaheuristic for solving continuous and discrete optimization problems. Our approach will be compared with this second approach in the experiments. Other works \cite{radovic2018evaluation, zolfani2018extended} respectively proposes and uses in transport field a method called SWARA that combine key performance indicators. However, experts must sort them according to their significance. Also, relative significance must be stated beforehand, that is, it must be determined how much an indicator is more important than the next indicator in the sorting. At this respect, experts in the waste collection are not able to sort them, and must less to establish beforehand a relative significance, then, this method has been discarded. Also, another work \cite{dezani2014optimizing} optimizes urban traffic flow using a genetic algorithm with Petri net analysis as fitness function. Their fitness function consists just in one key performance indicator, namely, the time, for whose estimation the Petri net is built. In our context, the time will also be a key performance indicator that will require to be estimated, but our proposal will involve performing a regression procedure. A Petri net cannot be applied in our context, since experts are required to build a Petri net. This task would involve establishing relationships among waste containers as the same way this work does among the roads using transitions. At this respect, experts do not report an existing relationship among containers. In fact, the original goal is to provide an optimized route planning, which means to establish the relationship among containers consisting in which container is collected next. They compare their approach with the lower distance. Hence, our approach will be compared with just consider the distance as fitness function. Another study \cite{nuortio2006improved} proposes a route planning and scheduling of waste collection and transport. The main drawback of this work is that it only uses one key performance indicator, namely, the time, which is estimated through a regression procedure from traveling distance. Our work goes further in the sense that experts reveal quite more issues involved in the route planning that just the time. For instance, they reveal the high influence of truck load and road kind as relevant aspects, so, this solution was discarded. All the same, our approach will be compared with the case of using the time key performance procedure as fitness function. 

The rest of the paper is organized as follows. Section~\ref{sec:1} describes some related work to this proposal. Section~\ref{sec:2} details the whole process of building a promising fitness function for capturing the waste collection planning expert knowledge, including the design. The description of the experiments and the discussion of the results are exposed in Section~\ref{sec:4}. Finally, Section~\ref{sec:5} draws some conclusions and proposes some lines of research for future work.
\section{Related Work} \label{sec:1}
There are a lot of works that take the expert and human information to generate promising approaches that capture the expert knowledge. For example, a work \cite{kubat1998machine} develops a model for detecting oil spill from expert information. Experts had to view images to detect suspicious regions and to classify them as positive or negative examples. Also, the experts provide the features, but the model obtained disregards many of them obtaining a good performance. Also, experts assess beef cattle as meat producers comparing animals with other partners \cite{bahamonde2004feature}. Some body dimensions obtaining by a 3D photo describe the beefs. The expert knowledge was captured in a ranking function that unifies the preferences of different experts. Similarly, activity rubric is adjusted from the ranking that lecturers made from student answers \cite{quevedo2009obtaining}. The goal is to determine the weight of each activity item using partial rankings made by lectures from several student answers to the activity.  Other study related to education \cite{garcia2018proposing} proposes a machine learning approach to obtain the most relevant factors that affect the employability and employment. To do that, some academics from different fields had to define the competencies that could be important for the employment using Spanish university data and student questionnaires. The aim of this study was to predict if a person will get employed or not and to extract the most relevant factors in order to know the best way for preparing the students for the labour market. Another work \cite{joachims2002optimizing} presents a strategy to optimize the search engine retrieval quality. The user knowledge about the document ranking related to a query is given to the search engine through the click through data. The document the user clicks on is preferred with regard to the rest of documents.  Support vector machines for ranking is fed with that information in order to generate a model able to rank the documents according to user preferences.

\section{A fitness function for capturing waste collection planning expert knowledge} \label{sec:2}
Before going in depth of obtaining a promising fitness function that captures the expert knowledge in waste collection planning, let us state some issues in order to better understand the current waste collection process. Actually, as commented before, there exists a set of routes locally planned by the experts from the successive council requests. The company has several garages in order to park the trucks. Each route begins in a garage and ends in the same garage.  The company also has transfer stations located in different places of the region where the trucks can make one or more stopovers in order to unload the waste if the cargo box is too full during a route. Also, each route has a group of collection geographic points that can include more than one container (of the same or of different capacity). If there is more than one container at the same collection point, the capacity at that collection point will be considered as the accumulated capacity of all the containers at that point. A truck and a human team are assigned to each route. The human teams are formed by collection workers and by drivers. The number of workers of a team depends on the route and none worker can exceed 38 work hours per week. Also, workers must not work more than 8 hours per day and the trucks almost have a limit time of being driven.
\\
A map of the container location together with the local designed routes allows concluding that the routes take the form of bouquets, since the majority have a central area with a high collection frequency and some branches with a low collection frequency. This is so because, the central area, which coincides with the settled area, is collected daily, but only one or two branches are included in the daily route. The experts have designed the routes per council (some routes cover more than one council if the areas are small) just using their experience and intuition causing the routes to have different duration. This is one of the pitfalls of the current waste collection planning, since there are some trucks and staff that overwork against others that underwork. 
\\
Some other issues include limitations in the container dimension in villages, the existence of vehicles that exceed the village weight and height limits or the allowed collection hour range at rush hours. Also, the possible delays in taking out the waste of some private companies because they have their container in a private enclosure or the adverse weather effects are other factors that make difficult the waste collection process. Besides, drivers might prefer to climb a road in worse conditions and to get down another road in better conditions than the opposite. Furthermore, the collection workers might prefer to collect the containers downhill than uphill. Another issue to keep in mind is the unloading time spent in transfer stations, which takes approximately two minutes. In addition to this, depending on the kind of truck, the running board, that is, the time to collect the waste can change. Moreover, the local governments determinate the amount of containers per surface and per population and decide the collection frequency of the container areas in terms of number of days. Owing to that, each container could have different frequency depending on the area. Furthermore, the collection frequencies of the containers may vary depending on the season of the year (Christmas time, summer, weekends...). It also happens that, for instance, in rural areas there are not time constraints to collect the waste and in some places container laterality is not an essential feature since sometimes the containers can be moved from one side of the road to the other. Other factor to take into account is the time that the driver takes from the garage to the first container on the route and from the last container to the garage.
\\

Given the complexity and casuistry of the waste collection process taking into account all the above-mentioned issues, defining a fitness function for evaluating a whole waste collection planning in order to be used in the future in a planning optimization algorithm to produce a global waste collection planning is clearly a non trivial task. Our proposal is to capture the expertise and knowledge of the experts fitting the problem in a preference learning framework. In this context, one expects to have the disposal of waste collection planning pairs for which the experts must express their preferences. This means that a preference learning algorithm is fed by what is called preference judgments. A preference judgement is a pair $(x,y)$ of route plannings such that the route planning $x$ is preferred to the route planning $y$ because the quality of $x$, $f(x)$, is higher than the quality of $y$, $f(y)$, that is, $f(x)\ge f(y)$, but the values of $f(x)$ and $f(y)$ are unknown. Therefore, a preference learning algorithm is fed by 
the set
\begin{equation}
\label{preferencejudgments}
S=\{(x,y): f(x)\ge f(y)\}
\end{equation}
Then, the preference learning statement consists in obtaining a preference or ranking function $f$ defined in the route planning space such that	 $f(x)\ge f(y)$ whenever $x$ is preferred to $y$, that is, when $(x,y)\in S$. This means that $f$ establishes an order or rank for the route plannings involved in $S$, that is, applying this function to the route plannings in $S$ one can order them. However, this learning goes further, since $f$ is a promising discovery because it is able to yield a rating for whatever route planning $x$ out of $S$.

Now, it is necessary to describe a route planning $x$ through features and show to the experts pairs of route plannings from which they decide which member of the pair is preferable taking into account their expertise and knowledge. The idea is to obtain expertise and knowledge from the experts from two different points of view. The first one is to discover the key indicators experts consider relevant in order to obtain a route planning description in form of features. The second one is to find out the preference judgment of the experts looking at the route plannings themselves, without taking into account explicitly the key indicators. Then, the ranking or preference function $f$ is induced from these information sources in a attempt to reproduce the expert decision knowledge. 
 
One of the difficulties that have arisen at this point is that the experts do not have the intellectual capacity of working in their minds with complete route plannings. They report serious difficulties in defining indicators for describing a whole route planning and also in deciding within a pair of route plannings which member of the pair is a better route planning. Hence, our proposal also includes making affordable the work of the experts in this sense. The idea at this respect consists of reducing the initial problem from the route planning space to the route space. In fact, up to now experts get use of designing routes in local areas, one isolated from the rest, so this simplification clearly makes the task more affordable for them. Then, pairs of routes rather than pairs of route plannings are presented to the experts, so they must decide which route is better in each pair of routes rather than decide which planning is better in each pair of route plannings. Also, our proposal establishes several key performance indicators to describe a route rather than a route planning taking into account the above-mentioned issues and factors about the waste collection process and the information coming from the experts. The indicators will be strategically designed in order to satisfy the additivity property. This feature makes them be suitable to further extend the solution from the route space to the route planning space. 

The discussion of the key performance indicators established is detailed in Section \ref{sec:KPIs}. One drawback arisen in designing these indicators is that travel time is not an available data, despite the experts state that it is a key issue to take into account. This is one of the obstacles that have been arisen in the design of key performance indicators. This work overcomes this drawback proposing a regression procedure in order to estimate the travel time distance from other available information. Up to now, only current local routes designed by the experts are available, but including the problem in a preference learning environment requires to have the disposal of other routes to built pairs from which experts must decide the member of each pair they prefer. In this sense, it arises the necessity of creating alternative routes able to design adequate route pairs to successfully allow acquiring the expert knowledge. This is another obstacle that has been arisen in the research of this paper. Our proposal copes with this drawback designing a greedy algorithm to create alternative routes for this purpose. The details are explained in Section \ref{sec:routes}. Finally, building a fitness function for evaluating a route through a preference learning process is exposed in Section \ref{sec:model}.

\subsection{Designing key performance indicators (KPIs) for a route} \label{sec:KPIs}
The KPIs for building a fitness function to assess the quality of a route are designed and exposed in this section. In some way, they are established in an attempt of gathering the most relevant factors that the collection experts consider as relevant when they need to design a waste collection route.
\\
In the following, the KPIs are discussed and designed according to the requirements of the experts. Let us notice that most of the factors the experts consider are related to geographic properties, then, a geographic information system will be required to compute the values of most of the KPIs.

\subsubsection{Travel distance}
\label{D}
The travel distance is one of the basic and straightforward KPI chosen by the experts. This measure is directly related to the fuel consumption and truck wear. In order to get its value, let consider a route whose initial and final garage is $g$ (let us remind that a requirement of the company is that each truck starts and ends in the same garage) and let be $C$ a set of waste collection points $\{p_1, p_2,..., p_n\}$ predefined by councils. Then, the travel distance ($D$) is obtained from several other smaller travel distances in the map, namely: 
\renewcommand{\labelenumi}{\roman{enumi})}
\begin{enumerate}
	\item Distance from the garage $g$ of the current route to the first waste collection point $p_1$ on the route, $D(g,p_1)$, and from the last waste collection point $p_n$ to the garage $g$, $D(g,p_n)$. 
	\item Distance between waste collection points that is computed as the sum of each distance between the current waste collection point $p_i$ and the next waste collection point $p_{i+1}$. 
	\begin{equation}
	D(p_1,p_n)=\sum_{i=1}^{n-1} D(p_i,p_{i+1})
	\end{equation}
\end{enumerate}
Hence, the travel distance of a route is the sum of all these distances. 
\begin{equation}
\label{4}
D(g,g)=D(g,p_1)+D(p_1,p_n)+D(p_n,g)
\end{equation}	

\subsubsection{Travel time}
\label{T}
The travel time is one of the crucial measure to take into account according to the experts and to other works \cite{dezani2014optimizing, nuortio2006improved}. Besides, experts justify it due to the existing timetable collection constraints and the maximum collection worker and driver shift duration. However, this measure is not easy to compute, as it seems. Even more, it is impossible to compute from the actual available information. In fact, only the time employed for the routes locally designed by the experts is at our disposal, so a mechanism to compute the travel time for a generic route must be established. The proposal to overcome this hurdle is to induce a regression function from the local information available through a learning process, for instance, as in case of \cite{nuortio2006improved}. Regression has been widely used in real applications as in case of prediction solar radiation over a meteorological station \cite{karasu2018solar}. In this process and given that a route is split into three stretches (see the discussion of travel distance), the route time computation is also split into the same parts in order to facilitate the learning process:

\begin{enumerate}
	\item Two time periods concern to the time spent in driving from the garage to the first waste collection point and from the last waste collection point to the garage. Each travel of both kinds is taken as a point for the regression process and both the travel distance and the travel time of the existing local routes are annotated. Then, a regression function $T_g$  is deduced for estimating the time spent in the two route stretches in which the garage is involved.
	\item The third time period regards to the time spent between collection points one each other. In this case, the distance between the first and the last collection point, the number of collection points and the travel time of the existing local routes are annotated to feed the regression system. Hence, another regression function $T_p$  is induced for predicting the travel time spent between collection points.
\end{enumerate}

Once both regression functions $T_g$ and $T_p$ are available, the prediction time spent in a certain route will be the sum of three parts: i) the one from the garage $g$ to the first collection point $p_1$, for which $T_g$ is applied, ii) the one from the first collection point $p_1$ to the last collection point $p_n$ for which $T_p$ is applied and finally iii) the one from the last collection point $p_n$ to the garage $g$, for which $T_g$ is applied again. Therefore, the total travel time of a route is the sum of all these time periods:

\begin{equation}
\label{5}
T(g,g)=T_g(g,p_1)+T_p(p_1,p_n)+T_g(p_n,g)
\end{equation}	
\\
Notice that it is possible that some of the predicted time periods may be greater than the actual ones. Hence, a correction factor will be applied for any route for which the regression functions are applied. This correction factor can only be computed from the available routes designed locally by the expert. For this purpose, the differences between real and predicted values are evaluated and the resulting sample is supposed to follow a normal distribution. The correction value will be the value for which the difference is positive at a certain significant level.

\vspace{2mm}
\subsubsection{Accumulated altitude}
\label{A}
Another relevant issue is the orography since a steep terrain complicates the access of the trucks to the collection points. Besides, it also affects the collection order, because it is obviously more costly for the trucks going uphill with a fuller load than going downhill with the same load, not only due to keeping the good condition of the trucks, otherwise also due to not rocket fuel consumption. One straightforward KPI in order to take into account this issue may be the accumulated altitude. One can also think of considering the slope percentage, but the experts reject this option, since it does not distinguish the best option between going uphill a long stretch with lower slope and going uphill a short stretch with higher slope. Then, and from a set of points of the route, $q_1,...,q_m$\footnote{These points can be collection points or not. In fact, they include the collection points and in addition quite more tracking points in order to get a better altitude estimation.}, their respective altitudes $A(q_1),...,A(q_m)$ are obtained and the accumulative altitude KPI for a route that begins at a garage $g$ and ends at the same garage is evaluated as follows

\begin{equation}
\label{5q}
A(g,g)=\sum_{i=1}^{m-1}\max(0; A(q_{i+1})-A(q_i))
\end{equation}

\subsubsection{Road type}
\label{BD}
The type of road is also a quite relevant factor, since driving along a motorway highly differs from driving along local roads. This fact is not only because of fuel consumption, otherwise it is also because of taking into account the driver comfort and the travel time spent.  According to the approval of the experts, the road types are classified according to their relevance (from major to minor) as motorway, dual carriageway, regional roads, local roads and others (square, street, path, avenue....). Then, and in order to consider the road type factor, three different KPIs are considered, namely, i) secondary road distance ($D_s$), which is the percentage of kilometres the truck travels along secondary roads (square, street, path, avenue and local roads) and ii) secondary and regional road distance ($D_r$), which is the percentage of kilometres the truck travels along secondary and regional roads. Hence, $D_s(g,g)$ and $D_r(g,g)$ are computed for a route (notice that the travel distance along main roads is the difference between the whole travel distance $D(g,g)$ and the sum of $D_s(g,g)$ and $D_r(g,g)$ and is not considered as a KPI, since it is a linear combination of the other KPIs, and hence it is redundant).
\\
\subsubsection{Truck load}
\label{L}
Another aspect that experts reveal great interest on is the truck load. They statistically prove that driving excessive kilometres with a heavy load highly affects the fuel consumption and truck wear.  In this direction, it is preferable to firstly drive to the farthest collection point of the route with the truck unload and to secondly return collecting the waste than taking the waste on the outbound journey, which means to carry with the whole waste loaded in the truck during all the return journey at least.
\\
Both fuel consumption and truck wear are almost impossible to estimate, then the truck load effect is not easy to summarize in a straightforward KPI. In fact, the KPI taken and able to compute that can represent the load truck will be a measure just related to fuel consumption (not even the fuel consumption itself) rather than the truck load itself. The proposal to establish a KPI for estimating this issue $E$ consists in measuring, in one sense, the effort of the truck to get the garage carrying the waste collected. For this purpose, let consider the distance from the current collection point $p_i$ to the garage $g$, that is, $D(p_{i}, g)$,  and let be $L(p_{i})$ the amount of waste the truck collects at the current point.  Then, the product of them
\begin{equation}
\label{LoadIntermedia}
E(p_i)=L(p_{i})\cdot D(p_{i}, g)
\end{equation}
can be an estimation of the effort of the truck to get the garage from the current waste collection point carrying the waste collected at this point. Hence, computing the sum of all $E(p_i)$ for all the collection points leads to a measure proportionally related to the fuel consumption and will define the KPI
\begin{equation}
\label{LoadTotal}
E(g,g)=\sum_{i=1}^{n}E(p_i)=\sum_{i=1}^{n}L(p_{i})\cdot D(p_{i}, g)
\end{equation}
A setting in order to effectively compare routes is to compute the average rather than just the sum, since the routes have, in general, different number of collection points. Hence, the KPI will be
\begin{equation}
\label{LoadTotalMedia}
E(g,g)=\frac{\sum_{i=1}^{n}E(p_i)}{n}=\frac{\sum_{i=1}^{n}L(p_{i})\cdot D(p_{i}, g)}{n}
\end{equation}

Furthermore, and unfortunately, neither the collected waste volume nor the collected waste weight is available data. Hence, one can think of estimating it, but the experts discarded this option, since they argued that it is highly probable that the noise that may be added in the estimation may not compensate the possible improvement obtained in exchange of taking a more accurate KPI. Consequently, the expert recommendation was to establish $L(p_i)=1$ for all the collection points and then to simplify the computation of the KPI for the current collection point as the remaining distance from the current point to the garage, that is, 
\begin{equation}
\label{LoadTotalMediaSimplificada}
E(g,g)=\frac{\sum_{i=1}^{n}E(p_i)}{n}=\frac{\sum_{i=1}^{n}D(p_{i}, g)}{n}
\end{equation}
As a final remark, we mention that the computation of this KPI assumes that the truck carries the waste until the garage, but this actually does not happen because the truck really empties the waste collected in a transfer station before getting the garage. Including this issue in the KPI highly complicates its computation, since, depending on the real last collection point, the transfer station may differ. 

\subsubsection{Final considerations}
As mentioned before, each route was split into three parts in order to compare the route pairs:
\begin{itemize}
	\item First stretch: from the garage to the first collection point
	\item Second stretch: between collection points 
	\item Third stretch: from the last collection point to the garage
\end{itemize}
Travel distance ($D$), travel time ($T$), accumulated altitude ($A$) and road type ($D_s$ and $D_r$) are computed not only for the whole route, otherwise, also for these three route stretches. However, only the KPI of the truck load $E$ is computed for the whole route. This means that a total of 21 KPIs are computed for each route. Table \ref{tab:kpis}
 summarizes all the KPIs computed.
\begin{table*}
	\centering
	\caption{The 21 KPIs computed for a route}
	\label{tab:kpis}
	\resizebox{10cm}{!} {
	\begin{tabular}{|r|c|c|c|c|c|c|}
		\hline
		& {\bf Travel} & {\bf Travel} & {\bf Accumulated} & {\bf Secondary } & {\bf Secondary and } & {\bf Truck} \\
		
		\multicolumn{1}{|c|}{\bf Stretch\textbackslash KPI} & {\bf distance} &  {\bf time} & {\bf altitude} & {\bf road distance} & {\bf regional road distance} & {\bf load} \\
		\hline
		{\bf 1st stretch} &{$D_1$} & {$T_1$} & {$A_1$} & {$D_{s,1}$} & {$D_{r,1}$} & {-} \\
		
		\hline
		{\bf{2nd stretch}} & {$D_2$} & {$T_2$} & {$A_2$} & {$D_{s,2}$} & {$D_{r,2}$} & {-} \\
		
		\hline
		{\bf 3rd stretch} & {$D_3$} & {$T_3$} & {$A_3$} & {$D_{s,3}$} & {$D_{r,3}$} & {-} \\
		
		\hline
		{\bf Complete route} & $D$ & $T$ & $A$ & $D_s$ & $D_r$& $E$ \\
		\hline
	\end{tabular}
}
\end{table*}

\subsection{Providing alternative routes}\label{sec:routes}
This section deals with the building of alternative routes for obtaining route pairs needed for being used in a preference learning framework that will provide a route fitness function. Our proposal consists of designing a greedy algorithm that builds adequate alternative routes. 
For each of the 64 existing routes, the coordinates of the garage and containers and the time limit for the route due to the workers and truck limit time is the information needed to feed the algorithm. According to the experts, the map distance $D$ is one of the key factors that makes great influence on the waste collection process cost and it is clue for the planning optimization. Since it is not available another fitness function that joins and combines the main factors exposed before (in fact, it is the goal of the paper), the map distance $D$ will be chosen as fitness measure to drive the greedy algorithm.
\\
For each existing route, the pseudocode of this process is shown in Algorithm \ref{algorithmRoute}. This algorithm starts with the truck assigned to this route leaving the garage to the nearest container of this route that has not been picked up. If adding this new container to the alternative route involves that the time of this alternative route is smaller than the workers and truck limit time, then, this container is added to that alternative route. In contrast, if it is higher, the algorithm finished and it is supposed that the truck goes back to its garage and the alternative route ends. 
\\

\begin{algorithm}
\caption{Algorithm for alternative waste collection routes}
\label{algorithmRoute}
\begin{algorithmic}[1]
    
\Require{$garage$, $P$ :   $garage$ is the garage where the route starts and ends. $P$ is the set of all waste collection points.} 
\Ensure{Route $R$ with all points}
\State  $list\_points \gets P$ \Comment{Init list with all the points of the route.}
\State $R.add(garage)$  \Comment{Add initial $garage$}
\State $p_{0} \gets select\_next\_point(garage,P)$   \Comment{Search next point from $garage$}
\State $R.add(p_{0})$
\State $list\_points.pop(p_{0})$ \Comment{Point $p_{0}$ is already selected}
\State $i =0$
\While {$len(list\_points)>0$}
\State $p_{i+1} \gets select\_next\_point(p_{i},list\_points)$   
\State $R.add(p_{i+1})$ \Comment{Add current point to $R$}
\State $list\_points.pop(p_{i+1})$
\State $i =i+1$

\EndWhile
\State $R.add(garage)$ \Comment{Add final $garage$}
\State{\textbf{return} $R$} 

\end{algorithmic}
\end{algorithm}

Let us now detail the way the closest container is taken each time. In order to find the truck with the smallest map distance to some container that has not been picked up yet, it is required to iterate for each pair of truck and container, and in addition, the external map service must be called at each iteration. This task is quite costly since calling the external map service would mean waiting a long time to get the answer. In addition, a very high number of requests would have to be made, which could cause the server to saturate. So it is quite interesting to find out a way of optimizing it. The proposal for performing an optimization for this process consists of considering the euclidean distance together with the map distance. Hence, the pseudocode of this optimization is shown in Algorithm \ref{algorithmPoint}.
\begin{algorithm}
\caption{Algorithm for obtaining the nearest map distance waste collection point}
\label{algorithmPoint}
\begin{algorithmic}[1]
\Require{$p_i$, $P$ :   $p_i$ is the current waste collection point. $P$ is the set of all waste collection points.} 
\Ensure{Next point $p_{i+1}$ from $p_i$}
\State  $list\_closest\_points \gets []$ \Comment{Init empty list to add next possible points.}
\State $p_{ednearest} \gets min\_euclidean\_distance(p_{i},P)$   \Comment{Nearest point of $P_{i}$ using euclidean distance ($p_{ednearest}$)}
\State $mapdistance \gets map\_distance(p_{i},p_{ednearest})$ \Comment{Get maximum limit of map distance}
\For {each point in $P$}
\If{ $euclidean\_distance(p_{i},point)<=mapdistance$}
\State{$list\_closest\_points.add(point)$} \Comment{Add it to closest points.}
\EndIf
\EndFor

\State {\textbf{return} $p_{i+1} \gets min\_map\_distance(list\_closest\_points)$ 
} 
\end{algorithmic}
\end{algorithm}

This way of obtaining the closest collection point allows searching among a limited number of neighbour points instead of performing a costly exhaustive search over the whole number of points that have not been collected yet. Besides, this process also guarantees to obtain the closest collection point because the Euclidean distance is always lower than or equal to the map distance. In this way, the optimum point is always obtained which leads to a notably reduction in the number of requests to be made to the external map service.

\subsection{Obtaining a model for ranking routes} \label{sec:model}
This section deals with the process of building a fitness function to assess the quality of a route through a preference learning framework. 

Once alternative routes are built, the KPIs exposed in Section \ref{sec:KPIs} are computed for both the existing routes locally optimized by the expert and the alternative routes built from them in Section \ref{sec:routes}. The KPI values will conform the route description $x=(D, T, A, D_s, D_r, E)$. Let us notice that all those KPIs are better as lower, since it is preferable a route $x$ with lower travel distance $D$, lower travel time $T$, lower accumulated altitude $A$, lower travel distance through secondary roads $D_s$, lower travel distance through secondary and regional roads $D_r$  and lower truck load $E$ than another route $y$ with higher KPI values. Then, the existing 64 routes can be split into three groups:

\begin{enumerate}[i)]
	\item The routes of the expert $x$ whose KPI values are all better or equal than the counterpart alternative route $y$.
	\item The alternative routes $y$ whose KPI values are all better or equal than the counterpart of the expert $x$.
	\item The routes $x$ with some KPI values better for the expert and the rest KPI values better for the alternative routes $y$.
\end{enumerate}
 
Among those 64 routes, just one route falls into the one that satisfies i) and just another one satisfies ii), whereas the rest of the routes belong to the iii) group. The pairs satisfying i) and ii) are directly considered to feed the preference learning algorithm, since the preference of the route is clear (the expert route in case of i) ($x\succ y$) and the alternative route in case of ii) ($y\succ x$). However, the preference for the routes of iii) is undefined and it is neither trivial nor easy to automatically provide a good decision. This is the reason why the experts were asked for deciding the best route in the case of this kind of pairs. The information provided to the experts to establish the preference judgments were just the routes placed on a map and not the KPIs that conform the route description.  They analysed the routes on the map and using their knowledge and expertise they revealed the preference judgment between the routes of each pair.\\

Once the pairs of routes are available with both the route description using the KPIs and the preference member established, the learning process takes place. As stated in Section \ref{sec:2}, the preference learning consists in obtaining a preference or ranking function $f$ defined in the route planning space such that	$f(x)$ is greater than $f(y)$ whenever $x$ is preferred to $y$. Let remember that $S=\{(x,y): f(x)\ge f(y)\}$ is the set of preference judgments, hence,
\begin{equation}
\label{preferencias1}
x\succ y \Rightarrow f(x)>f(y) \Rightarrow f(x)-f(y)> 0
\end{equation}
and if $f$ is linear, then:
\begin{equation}
\label{preferencias2}
f(x)-f(y) = f(x-y)
\end{equation}
and therefore 
\begin{equation}
\label{nueva}
x\succ y \Rightarrow f(x-y)>0
\end{equation}
Hence, the ranking function $f$ can be learned using a binary classification algorithm able to separate the class according to the sign returned. Consequently, once the KPIs are calculated, each pair of routes is transformed into a pair of examples, where the classes are +1 and -1. If $x\succ y$, the pair of examples are
\begin{eqnarray}
\label{preferencias3}
T=\{(x-y;+1), (y-x; -1): (x, y)\in S\} 
\end{eqnarray}
The linear function induced by the learning process $f$ passes through the origin of coordinates and is thus defined by
\begin{equation}
\label{fl}
f(z)=<w,z>=\sum_{i=1}^{21}w_iz_j
\end{equation}
where $w$ is the weight vector and $<w,z>$ is the scalar product of $w$ and $z$. The score $f(z)$ will be the assessment of $z$ in the sense that $f(z)$  will be taken to predict preferences between the route $z$ and other routes. Also, $w$ will be the director vector of the assessment hyperplane.

Let us notice that it is a good choice to assume that $f$ is linear, since, as commented before, the values of the KPIs are better (preferable) as lower, that is, they are monotone. 
Under this paradigm, the problem can be solved using a binary classification algorithm \cite{herbrich2000large}. However, not all binary classification algorithms are suitable in this context. Firstly, the algorithm must obtain a linear model. Secondly, the algorithm must be one among those that produces a numerical value. This numerical value is commonly taken afterwards for performing the classification into $+1$ or $-1$ classes. In fact, this value is typically higher as the algorithm is more reliable of an instance to belong to class $+1$ and it is this property what makes this value be suitable for providing a ranking. Hence, the afterwards classification will be ignored in this preference learning framework. Two promising options have been found that fit these criteria, namely, Support Vector Machines (SVM) \cite{PlattProbabilisticOutputs1999,Wu2004} with linear kernel and Logistic Regression (LR) \cite{liblinear,HsiaCL18,LinWK07}. Both algorithms have been widely and recently used in real applications, for instance, for financial time series forecasting \cite{altan2019effect} or for predicting proportions of complex blends in food products \cite{carrasco2018local} in case of SVM and for aircraft engine degradation prognostics \cite{lu2019aircraft} or for outdoor thermal sensation comfort range \cite{xie2019outdoor} in case of LR. Although there exist much more algorithms for binary classification, as $k$-nearest neighbour, neural networks or decision trees, they do not provide a linear model or they do not produce a numerical value related to the classification.

\subsection{Assessing KPIs through feature selection}\label{sec:fs}
This section performs a study about the quality of the KPIs in the performance of route ranking. Although the KPIs have been carefully and exhaustively designed under the criteria of the experts, let us go in depth and analyse if effectively all of them are necessary and suitable to obtain a promising route ranking. In fact, the experts have serious doubts about their influence. For this purpose, we have checked several ways of selecting KPIs. The experts proposed the first one. They were unsure about if the KPIs must be taking globally for the whole route or by stretches depending if the stretch involves collection points or not. Also, they had special interest in checking the load truck influence. In this sense, they have proffered the following groups of KPI features to analyse:
\begin{enumerate}
\item Whole route KPIs: $D$, $T$, $A$, $D_s$ and $D_r$.
\item Whole route KPIs and truck load: $D$, $T$, $A$, $D_s$, $D_r$ and $E$.
\item KPIs by stretches: $D_i$, $T_i$, $A_i$, $D_{s,i}$ and $D_{r,i}$ for $i=1, 2, 3$.  
\item KPIs by stretches and truck load: $D_i$, $T_i$, $A_i$, $D_{s,i}$, $D_{r,i}$ for $i=1, 2, 3$ and $E$.
\item KPIs by stretches and whole route KPIs: $D$, $T$, $A$, $D_s$, $D_r$ and $D_i$, $T_i$, $A_i$, $D_{s,i}$ and $D_{r,i}$ for $i=1, 2, 3$.  
\item KPIs by stretches, whole route KPIs and truck load: $D$, $T$, $A$, $D_s$, $D_r$ and $D_i$, $T_i$, $A_i$, $D_{s,i}$, $D_{r,i}$ for $i=1, 2, 3$ and $E$.  
\end{enumerate} 
Automatic feature selection was also adopted in order to compare with the groups of features taken by the experts. Feature selection is a widely practice \cite{karasu2019recognition} to improve the performance of the machine learning techniques, since it removes redundancy and noise included in the data. The idea here is to take an adequate feature selection technique suitable for algorithms that produce linear models and that get good performance in case of few features. A promising technique for this purpose is Recursive Feature Elimination (RFE) \cite{guyon2002gene}. This method is a greedy algorithm that repeatedly builds a model removing the worst feature according to assign weights to features, commonly the coefficients of a linear model. It is considered a wrapper, since a machine learning algorithm is used in the core of the method to build the model. Then, it successively repeats the process with the remaining features until the desired number of features to select is eventually reached, although the algorithm is typically tuned for wording until no more features are available. Finally, the features are ranked in the same order they have been removed. The pseudocode of the method is shown in Algorithm \ref{algorithmRFE}.
\begin{algorithm}
\caption{Recursive Feature Elimination (RFE) algorithm}
\label{algorithmRFE}
\begin{algorithmic}[1]

\Require{$Data$:   Dataset T (see Equation (\ref{preferencias3})) with preference judgments that contains $v^*$} 
\Ensure{Ranked list of indicators $(z_1,...,z_{v^*})$ order by their relevance.}
\State  $tuning\_parameters(model)$
\State  $train(model)$

\State  $v \gets v^*$

\While {$v\geq2$}
\State $model_v \gets model$ with the optimized tuning parameters for $v$ and $Data$
\State{$w_v \gets calculate\_weight\_vector(model_v)$} \Comment{$(w_{v1},...,w_{vv})$}
\State  $rank\_criteria \gets (w_{v1}^2,...,w_{vv}^2)$
\State  $min\_rank\_criteria \gets  min(rank\_criteria)$ \Comment{Lowest value}

\State  $Remove(min\_rank\_criteria,Data)$ \Comment{Remove from $Data$}
\State  $z_v \gets min\_rank\_criteria$

\State  $v \gets v-1$
\EndWhile
\State  $z_1 \gets $ variable in $Data \notin (z_2,...,z_{v^*})$
\State {\textbf{return} $(z_1,...,z_{v^*})$ 
} 

\end{algorithmic}
\end{algorithm} 

The strategy that follows this algorithm makes it be of linear order with regard to the number of features, in contrast to other algorithms, as Backwards Elimination (BE) or Forward Selection (FS), which are of quadratic order with regard to the number of features. 
The RFE method requires to be used together with algorithms that produce linear models, which is a condition above-established in the preference learning framework and the main reason why SVM and LR are chosen.  Besides, RFE has been successfully and recently applied to other real applications, as for instance, for Alzheimer’s disease diagnosis  \cite{richhariya2020diagnosis}, for credit card fraud detection \cite{rtayli2020enhanced} or for heat-resistant steel running state evaluation \cite{huang2019hybrid}.
 \section{Experiments}\label{sec:4}
This section deals with the experiments carried out in this work, but before establishing the parameter settings and displaying and discussing the experiment results, let us state the geographic information systems for obtaining the KPIs (see Section \ref{sec:gis}). Concerning the experiments, two different types are exposed, namely, i) experiments to estimate the travel time KPI through a regression process (see Section \ref{sec:timeestimation}) and ii) experiments through a preference learning process to provide the fitness function to evaluate the quality of a waste collection route that, as it was previously mentioned, it will be also taken to evaluate the quality of a waste collection planning  (see Section \ref{sec:functiondesign}); this last kind of experiments also includes the feature selection process. 
\subsection{Geographic information systems for obtaining the KPIs}\label{sec:gis}
The KPIs are computed using a geographic information system called Open Source Routing Machine (OSRM)\footnote{\url{https://map.project-osrm.org/}}, which is based on OpenStreetMap (OSM)\footnote{\url{http://www.openstreetmap.org}}. There are a variety of applications or projects that have been developed using OSM\footnote{Thunderforest project (\url{http://www.thunderforest.com/}), the Open Topo Map (\url{https://opentopomap.org/}),  the MTBmap (\url{https://openmtbmap.org/es/}) or the Waymarked Trails project (\url{http://www.waymarkedtrails.org/})} and OSRM\footnote{Cycle.Travel (\url{http://cycle.travel/}), I Bike Cph (\url{https://www.ibikecph.dk/en}) and MAPS.ME (\url{https://maps.me/})}. 
\\
OSMR does not include altitude data and the information may be quite incomplete in low population density areas, since it depends on the response of users. Despite that, there are several advantages of OSRM that makes it be a good choice, namely, it is free, it is possible to find very extensive documentation, it is optimized for the transport mode used to move between locations and it allows to choose the type of truck to carry out the routes. 
\\
Other alternatives to OSMR are Google Maps\footnote{(\url{https://www.google.com/maps})} that incorporates elevation data and very complete information, GraphHopper\footnote{(\url{https://www.graphhopper.com/})} that has a route optimization API and elevation data or Mapbox\footnote{(\url{https://www.mapbox.com/})} that is equipped with a traffic-oriented routing module or Mapzen Valhalla\footnote{(\url{https://mapzen.com/})} that admits trips by car, on foot, by bicycle and by public transport, calculates the most efficient way to visit multiple destinations and has elevation data even for bicycle routes. However, all of them present the disadvantage of being not free for huge amount of request to their servers, something that is crucial for us. 
\\
The alternative to provide altitude data needed to compute the accumulative altitude $A$ that we propose to use in this paper is the global elevation model called GMTED2010\footnote{https://lta.cr.usgs.gov/GMTED2010}, since the number of requests is not limited and it is able to cope with waste collection vehicles. This software was developed by the U.S. Geological Survey (USGS) and the National Geospatial-Intelligence Agency (NGA). In this sense, OSRM provides a set of points of a stretch and, then, the GMTED returns the altitude of these points.
\begin{table}	
\caption{RMSE and RMSE SD for $T_g$ and $T_p$ regression models}
\label{tab:tablatraveltime}
	\centering
	\resizebox{9cm}{!} {
		\begin{tabular}{l | c c | c c }
		       \hline
		       & \multicolumn{2}{c|}{$T_g$ model} & \multicolumn{2}{c}{$T_p$ model}\\
			\hline 
			\textbf{Algorithm} & \textbf{RMSE} & \textbf{RMSE SD} & \textbf{RMSE} & \textbf{RMSE SD}\\ 
			\hline
			LiR & 489.65 & 82.18  & 525.72 & 163.41\\ 
			BC & 530.91 & 83.69 & 693.10
			& 285.93
				\\ 
			BT & 498.67 & 75.34 & 665.98 & 273.68 
			\\ 	
			RF & 542.87 & 78.55 & 521.35 & 156.71
			\\ 
			BRR & \textbf{489.50} & 83.00 & \textbf{520.70} & 167.14\\ 
			\hline 
		\end{tabular}
	}
\end{table}
\subsection{Travel time estimation}\label{sec:timeestimation}
Two models have been created to estimate the time spent by a truck during the route. The first model will estimate the time employed by the truck in driving from a garage to the first waste collection point that will be the same that will estimate the time employed from the last waste collection point to the garage. The second model will estimate the time employed between containers. 
\\
Both models were created using the library Caret4 (short for classification and regression training) in R. This library includes several algorithms for the regression and classification tasks. The use of one algorithm or another might influence the purpose of this project, changing the routes and their times. Consequently, we perform an exhaustive study in order to the get good travel time estimation. Particularly, we carried out experiments using linear regression (LR), bagged CART (BC), Boosted Tree (BT), Random Forest (RF) and Bayesian Ridge Regression (BRR) methods.
\\
For all these methods, a cross validation was performed with 3, 5 or 10 folds with 3 repetitions. Only BT and RF require parameters that were tuned and decided to be $maxdepth = 2$ and $mstop = 50$ for BT and $mtry=1$ for RF.
\\
Table \ref{tab:tablatraveltime} displays the root mean square error (RMSE) and its standard deviation (RMSE SD) for the above-mentioned methods for estimating the travel time $T_g$ (from garage to waste collection point and vice versa) and $T_p$ (between waste collection points). The method that offers the best performance for both $T_g$ and $T_p$ is BRR, and it will be the method we will choose to estimate the travel time. However, LiR might be also a good alternative for estimating both $T_g$ and $T_p$ taking into account that i) the RMSE of LiR is slightly lower than for BRR, ii) the RMSE SD values of LiR are quite similar to that of BRR and iii) the magnitude of RMSE SD with regard to the RMSE values. The same happens for BT, but only in case of estimating $T_g$, and for RF, but only in case of estimating $T_p$. It is quite remarkable that the RMSE SD values in case of estimating $T_g$ are quite lower than those values in case of estimating $T_p$, maybe because the casuistry is clearly more complex in case of the travels between collection points than in case of the garage to the first collection point and of the last collection point to the garage.
\begin{figure}[t]
\centering
\includegraphics[width=1\textwidth]{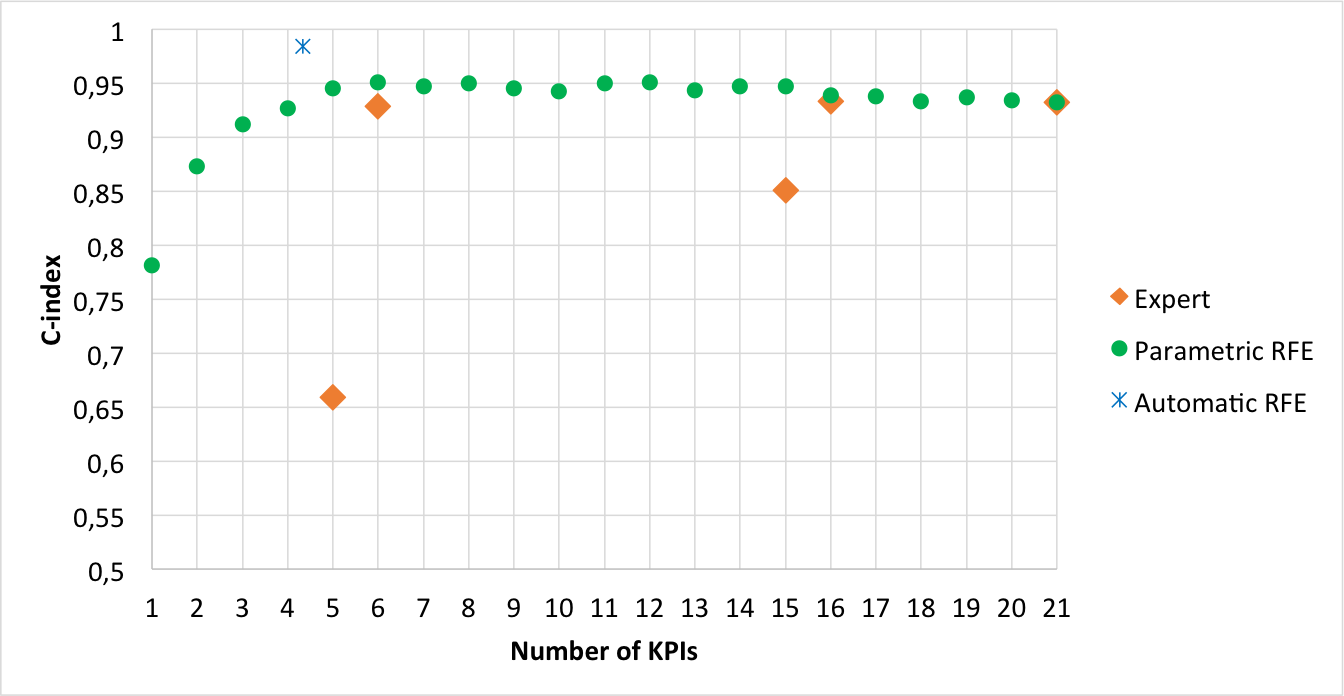}
\caption{$C-$index for expert feature selection, parametric RFE and automatic RFE when SVM is taken as an algorithm that produces linear models}
\label{fig:graficoSVMRFE}
\end{figure}
\begin{figure}[t]
\centering
\includegraphics[width=1\textwidth]{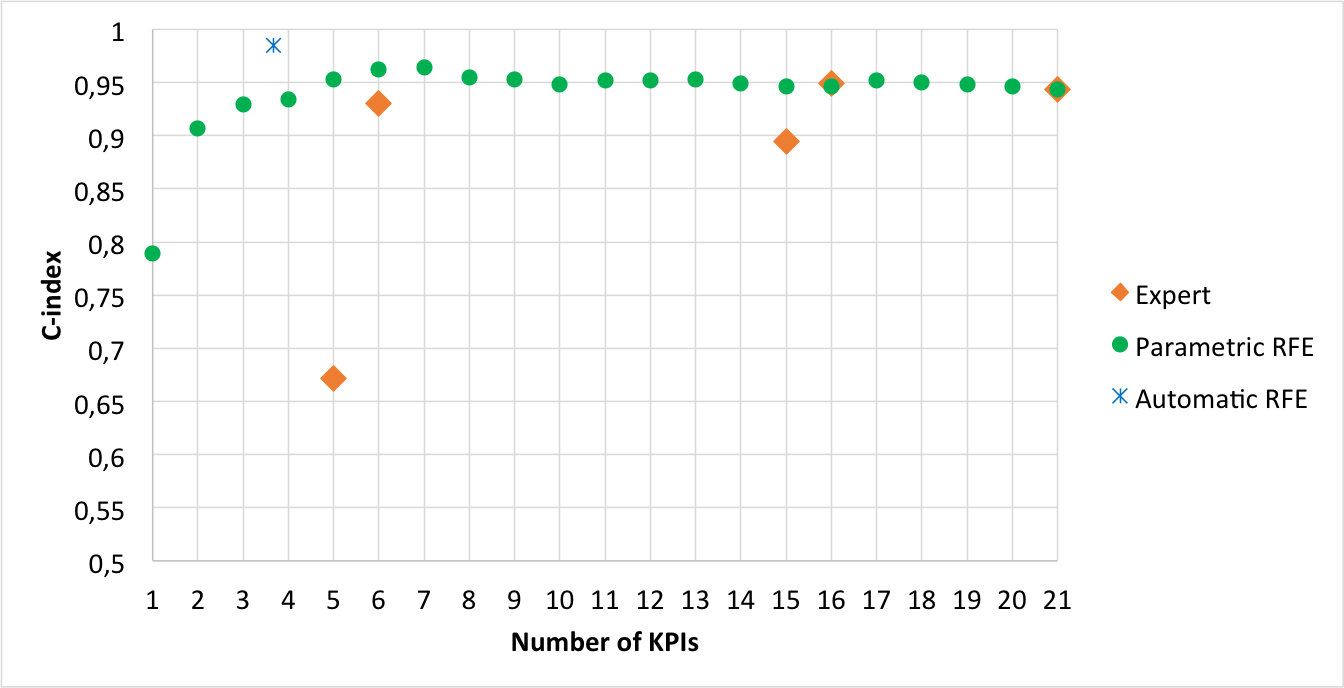}
\caption{$C-$index for expert feature selection, parametric RFE and automatic RFE when LR is taken as an algorithm that produces linear models}
\label{fig:graficoLRRFE}
\end{figure}
\subsection{Fitness function for ranking routes}\label{sec:functiondesign}
Before presenting and discussing the results of the fitness function design, let us discuss the experimental settings, the methods for both providing a ranking and feature selection and the performance measure employed.
\subsubsection{Experimental settings}
SVM \cite{PlattProbabilisticOutputs1999,Wu2004} with linear kernel and and LR \cite{liblinear,HsiaCL18,LinWK07} were chosen to provide a fitness function for ranking routes, since, as mentioned in Section \ref{sec:model},  they are promising algorithms that produce linear models and that provide a numerical value to make a decision about the class value that are suitable in the preference learning context.

A comparison with other approaches discussed in Section \ref{sec:0} was performed. On one hand, experiments were carried out just considering a unique KPI \cite{dezani2014optimizing,nuortio2006improved}. But, our study goes further and not only travel time ($T$) and travel distance ($D$) are considered isolated from the rest, otherwise experiments were also performed for the accumulated altitude ($A$), road type ($D_s$ and $D_r$) and truck load ($E$). On the other hand, and following the work \cite{leontiadis2017good} that combines key performance indicators from mobile networks, experiments with the optimization algorithm called particle swarm optimization \cite{blackwell2007particle} were also carried out. Each particle is represented by $21$ values, which correspond to the weights of the $21$ KPIs. The fitness function considered was the mean $C-$index of the ranks of each preference judgment. The parameter values were set according to \cite{leontiadis2017good}. Particularly, the number of iterations was fixed to $1000$ and the number of particles $10$, $50$, $100$ and $1000$. 

Concerning feature selection, in addition to the manual feature selection provided by the experts, the proposal was to use RFE \cite{guyon2002gene} (see Section \ref{sec:fs}). This method has been checked using two different manners. The first modality (that we will call parametric RFE) consists of setting beforehand the desired number of features to select, which ranges from 1 to 21 whereas the second modality (that we will call automatic RFE) involves letting the method to provide the best set of features. 

\begin{table}[htb]	
\caption{$C-$index when just a unique KPI is taken for fitness function}
\label{tab:resultuniquekpi}
\centering
\begin{tabular}{c c c c c c}
\hline
$D$ & $T$ & $A$ & $D_s$ & $D_r$ & $E$\\
\hline
57.81 & 60.94 & 50.00 & 53.13 & 54.69 &71.88\\
\hline 
\end{tabular}
\end{table}
			
\begin{table}[htb]
\caption{$C-$index for particle swarm optimization for different number of particles}
\label{tab:resultpso}
	\centering
		\begin{tabular}{l | c c c c}
		       \hline
Particles & 10 & 50 & 100 & 1000\\
\hline
$C-$index & 81.51 & 86.57 & 87.11 & 89.65\\
			\hline 
		\end{tabular}
\end{table}
			
\begin{table}[htb]
\caption{Features selected by the parametric and automatic RFE when SVM and LR is taken as an algorithm that produces linear models}
\label{tab:featuresselected}
\centering
	\resizebox{14cm}{!} {
		\begin{tabular}{ll}
	\hline
		\multicolumn{2}{c}{Parametric RFE}\\	
		       \hline
		        \textbf{SVM ($6$ features)} &  \textbf{LR ($7$ features)}\\
		       \hline
		       \textbf{truck load ($E$)}	&\textbf{truck load ($E$)} \\
		       altitude ($A$) &\textbf{1st stretch distance ($D_1$)} \\
			\textbf{1st stretch distance ($D_1$)} &\textbf{2nd stretch distance ($D_2$)}\\
			\textbf{2nd stretch distance ($D_2$)} &3rd stretch distance ($D_3$)\\
			2nd stretch secondary and regional road distance ($D_{r,2}$) &3rd stretch time ($T_3$)\\
			\textbf{3rd stretch altitude ($A_3$)}&\textbf{3rd stretch altitude ($A_3$)}\\
			&3rd stretch secondary and regional road distance ($D_{r,3}$)\\
		       \hline 
		\multicolumn{2}{c}{Automatic RFE}\\
		 \hline
		  \textbf{SVM ($6$ features)} &  \textbf{LR ($8$ features)}\\
		       \hline
		  \textbf{truck load ($E$)} & \textbf{truck load($E$)}\\
\textbf{1st stretch distance ($D_1$)}& \textbf{2nd stretch secondary and regional road distance ($D_{r,2}$)}\\
1st stretch secondary and regional road distance ($D_{r,1}$) &altitude ($A$)\\
1st stretch time ($T_1$)&secondary and regional road distance ($D_r$)\\
\textbf{2nd stretch secondary and regional road distance ($D_{r,2}$)} &3rd stretch time ($T_3$)\\
\textbf{3rd stretch secondary road distance ($D_{s,3}$)}&\textbf{3rd stretch secondary road distance ($D_{s,3}$)}\\
&2nd stretch distance ($D_2$)\\
&\textbf{1st stretch distance ($D_1$)}\\
\hline 
		\end{tabular}
	}
\end{table}
\begin{figure}[h]
\centering
\includegraphics[width=.9\textwidth]{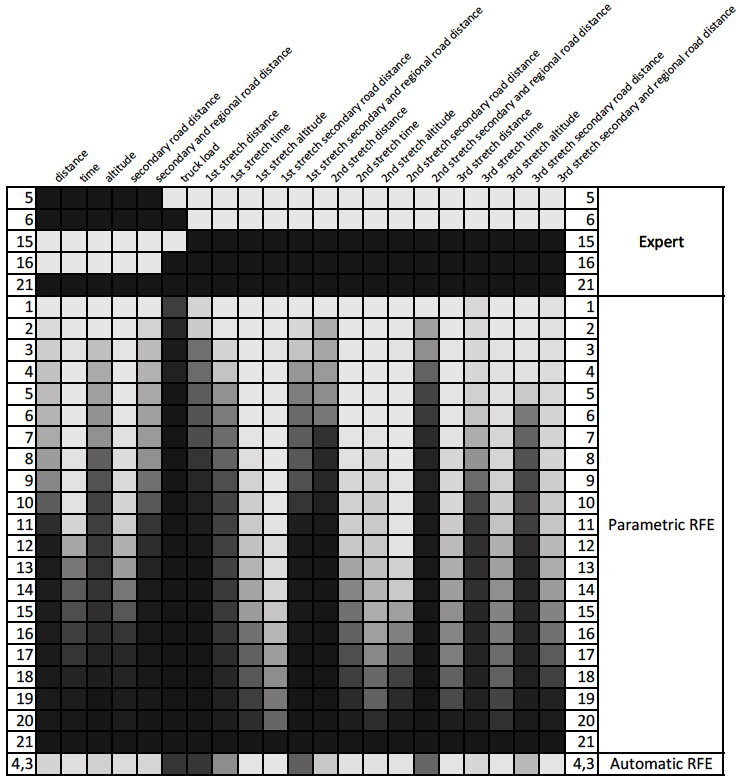}
\caption{$C-$index for expert feature selection, parametric RFE and automatic RFE when SVM is taken as an algorithm that produces linear models}
\label{fig:mapaSVM}
\end{figure}
\begin{figure}[h]
\centering
\includegraphics[width=.9\textwidth]{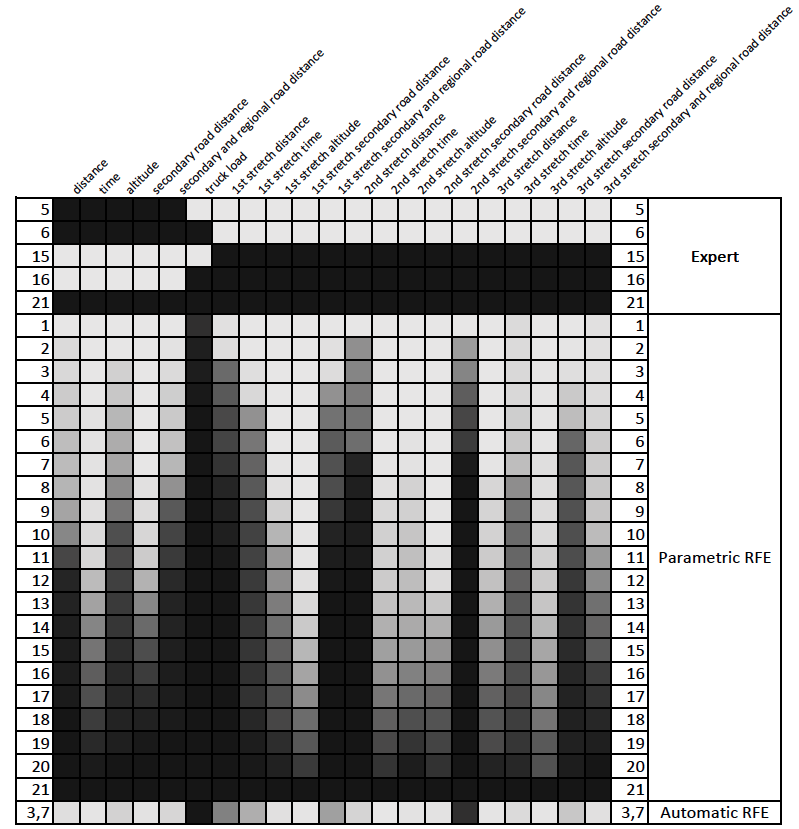}
\caption{$C-$index for expert feature selection, parametric RFE and automatic RFE when LR is taken as an algorithm that produces linear models}
\label{fig:mapaLR}
\end{figure}
The $C-$index is the evaluation measure taken.  It is an estimator of the concordance probability \cite{GonenHeller2005} typically used in statistics and equivalent to the pairwise ranking error for ordinal regression \cite{herbrich2000large} that has been commonly used in multipartite ranking as an estimation of the probability that a randomly chosen pair of instances from different classes is ranked correctly \cite{furnkranz2009binary} as a straight-forward generalization of AUC measure. The $C-$index was estimated though a cross-validation of 10 folds and 10 repetitions.
\subsubsection{Experimental results}
This section deals with the results reached in the experiments. Firstly, the results from the existing approaches are discussed. Then, the results provided by the proposed approach are exhaustively analysed from different points of view. 

Tables \ref{tab:resultuniquekpi} and \ref{tab:resultpso} display the results of the existing approaches. Particularly, Table \ref{tab:resultuniquekpi} shows the $C-$index when just a unique KPI is taken as fitness function, whereas Table \ref{tab:resultpso} displays the $C-$index for the particle swarm optimization algorithm with different number of particles. The conclusions are the following: 
\begin{enumerate}
\item Between distance $D$ and time $T$, one can observe that the fitness function consisting just of the time indicator produces the best performance. This results agrees with the conclusions reached in other approaches \cite{dezani2014optimizing,nuortio2006improved}. However, the truck load $E$ yields the best performance. So far, and to best of our knowledge, none work has considered the truck load as an indicator, in spite of experts in the field highly expected to make great influence on the route planning assessment.
\item The particle swarm optimization provides quite best performance with regard to the method that just takes a unique KPI. This fact confirms the hypothesis that combining several issues for providing a fitness function improve the performance of the route planning assessment. The performances improve as the number of particles increases, but in any case it does not reach the $90\%$ of $C-$index.
\end{enumerate}

Figures \ref{fig:graficoSVMRFE} and \ref{fig:graficoLRRFE} respectively display the $C-$index when SVM and LR is taken for providing a linear model together with expert feature selection, parametric RFE and automatic RFE. Clearly, fitting the problem into a preference framework overtakes the other existing approaches in the literature. 

Focusing now exclusively on the method proposed in this paper, one can extract the following conclusions from those figures:
\begin{enumerate}
\item The load truck is a crucial KPI to rank the routes (see the diamonds when $6$ versus $5$ features are selected and the same when $15$ versus $16$ features are selected). This is so, since the performance clearly improves both in the case of taking the whole route KPIs and in case of taking the KPIs split in stretches, especially in the former case. 
\item It seems that there are hardly differences between taking the KPIs for the whole route and taking the KPIs by stretches (compare the diamonds at $6$ features versus $16$ features).
\item Automatic RFE gives the overall best performance (see the cross in both figures), whose performance is around $98\%$.
\item There clearly exists a high redundancy among the KPIs, as experts expected. This is so because both parametric and automatic RFE reaches the best performance when few features are selected. Particularly, the parametric RFE offers the best results when it takes $6$ (for SMV) or $7$ (for LR) out of $21$ features and automatic RFE does it taking around (an average along folds and repetitions) $4$ of the $21$ features. 
\end{enumerate}

Let now analyse the more promising KPIs, that is, the KPIs that yield the best performance. Table \ref{tab:featuresselected} shows the features selected by the parametric RFE when they reach the best performance and also by the automatic RFE. The main conclusions of this analysis are:
\begin{enumerate}
\item In regard to parametric RFE, the selected features differ if one uses SVM or LR. They select $4$ common features (bolded in Table \ref{tab:featuresselected}) out of $6$ for SVM and out of $7$ for LR, which is respectively about the $66\%$ and $57\%$ of the features selected. These common features are truck load ($E$), the first and second stretch distances ($D_1$ and $D_2$) and the third stretch altitude ($A_3$). 
\item Concerning automatic RFE, which reaches the overall best performance, it selects $6$ out of $21$ features using SVM to get the linear model, whereas it selects $8$ out of $21$ when it uses LR. They select $4$ features in common (bolded in Table \ref{tab:featuresselected}), namely truck load ($E$), the first stretch distance ($D_1$), the second stretch secondary and regional road distance ($D_{r,2}$) and the third stretch secondary distance ($D_{s,3}$). These common features respectively are about the $66\%$ for SVM and $50\%$ for LR of the features selected. 
\item As seen, the truck load ($E$) is a clue feature in this process, as it was expected. Also, the distance in the three stretches are highly relevant, although depending on the stretch, this distance is taken for the whole stretch (in case of the first stretch), only for secondary and regional part of the stretch (in case of the second stretch) and for the secondary road stretch (in case of the third stretch). 
\item Another conclusion is that splitting the route in stretches seems to be a good choice, since the method does not tend to select features of the whole route. In fact, none whole route feature is taken in case of SVM. 
\end{enumerate}

Figures \ref{fig:mapaSVM} and \ref{fig:mapaLR} respectively display a heat map that represents the frequency whit which each feature is selected by parametric (in the middle) and automatic (at the bottom) RFE in the cross-validation folds (taking also into account the repetitions) of the cross validation.  They also include the manual selection proposed by the experts as a baseline reference. The following issues summarizes the conclusions: 
\begin{enumerate}
\item At sight, the most frequently feature is clearly the truck load ($E$). Other features of the whole router that use to appear are distance ($D$), altitude ($A$) and secondary and regional road distance ($D_{r}$). 
\item Focusing on stretches, distance ($D_1$), time ($T_1$) and secondary and regional road distance ($D_{s,1}$) predominate in the first stretch, distance ($D_2$) and secondary and regional road distance ($D_{r,2}$) do so in the second stretch and time ($T_3$) and secondary road distance ($D_{s,3}$) do so in the third stretch. 
\item Viewing the map horizontally, one can find out redundancy in the features. Let see, for instance, two cases as an illustration:
\begin{enumerate}
\item Case of parametric RFE when it selects $2$ features. In this case, second stretch distance ($D_2$) and second stretch secondary and regional road distance ($D_{r,2}$) seems to appear as alternative one from another together with the truck load. This fact is often repeated when more than $2$ features are selected. 
\item Case of $7$ features. In this case, about $11$ features used to be selected. Truck load ($E$), second stretch distance ($D_2$) and second stretch secondary and regional road distance ($D_{r,2}$) seem to be almost always selected, whereas the $4$ remaining alternate one from each other. Besides, this behaviour is common both for SVM and LR. 
\end{enumerate}
Although this redundancy seemed to be latent and now it is visually displayed, it does not have caused surprise among the experts. They always were aware of existing relationships among the entire feature they proposed, although they recognize they are unknown. 
\end{enumerate}
\section{Conclusions and future work}\label{sec:5}
This paper takes the first step in establishing an automatic and global optimization on the waste collection process of the company called \textit{Consorcio para la Gestión de los Residuos Sólidos de Asturias} (COGERSA). This step consists of designing a fitness function to assess a candidate route planning in order to use it in planning optimization algorithm for providing a global optimal route planning. So far, the experts manually designed the route planning focusing on particular local areas with a relative reduced number of waste containers whenever and under council demand. 
The route planning complexity due to the existing resources and constraints even makes the task of defining a fitness function not be affordable by the experts. This paper provides a fitness function from the expert knowledge and stating its design in a preference learning framework. The process is simplified to design a fitness function that assesses a route rather than a route planning again due to complexity. Fortunately, this simplification was easy to carry out, since expert did not mention specific indicators of a planning route, maybe because so far they planned routes one isolated from the rest. 
The idea under preference learning is to obtain expertise and knowledge from the experts from two different points of view. The first one is to discover the key indicators experts consider relevant in order to obtain a route description in form of features. The second one is to find out the preference judgment of the experts looking at the routes themselves, without taking into account explicitly the key indicators. Then, a ranking or preference function is induced from these information sources in an attempt to reproduce the expert decision knowledge, which will be the fitness function. For this purpose, this paper has exhaustively and carefully designed several key performance indicators according to the experts. These indicators were established for describing a route, but all of them have the additivity property, hence, they are straightforward extended to a route planning. Also, experts expect redundancy among such indicators, and for this reason, a feature selection analysis was included in this paper in order to check in what extent this hypothesis holds. The conclusion of the study confirms the existence of high redundancy among the indicators, since the best performance is reached when $6$ or $8$ out of $21$ indicators are taken. Truck load seems to be a clue indicator together with the distance travelled, especially if such distance belongs to no main roads. The proposed approach has been compared to other existing methods in the literature, that just consider a unique indicator as fitness function or combine them through a particle swarm optimization. The conclusion is that the method proposed in this paper clearly outperforms the existing approaches, since the performance goes from $72\%$ for a unique indicator fitness function or $90\%$ for particle swarm optimization to $98\%$ of the proposed preference learning approach.
\\
As future work, the planning consists of including the fitness function in planning optimization algorithms. Besides, and given the existing redundancy among the key performance indicators, we plan to study more in depth those that have been shown to be clue in this study. For instance, truck load is one of the most relevant indicators. However, the way experts have found for computing it is quite gross. Hence, obtaining an alternative way of estimating it in a more accurate way may lead to an improvement. Also, the distance travelled along different kinds of roads seems to determine the quality of the route. In this sense, it may be promising to polish the split of the roads in different kinds, taking into account not only the official type of the route, otherwise the state, traffic and other environment issues. Finally, and since the original goal was to optimize a route planning it would be the great interest to consider key performance indicators of route planning, for instance, the number of routes, highly related with the human and material resources.
\section*{Acknowledgments}\label{sec:Acknowledgments}
This research has been partially supported by the Spanish Ministerio de Ciencia e Innovación through the grant PID2019-110742RB-I00) and by the RDI project Smart Waste Collection (SWC), developed by the consortium COGERSA, SADIM S.A., S.M.E - grupo HUNOSA and ABAMobile and partially supported by the Instituto de Desarrollo Económico del Principado de Asturias (IDEPA) through the grants IDE/2015/000863, IDE/2015/000864 and IDE/2015/000865.

\bibliographystyle{elsarticle-harv}
\bibliography{mybib}

\end{document}